\documentclass[runningheads,a4paper] {llncs}

\usepackage{amssymb}
\usepackage{amsbsy}
\usepackage{graphicx}
\usepackage{tabularx}
\usepackage{url}
\usepackage{pbox}
\usepackage{multirow}
\usepackage{booktabs}
\usepackage{threeparttable}
\usepackage{floatrow}
\usepackage{bm}
\usepackage{marvosym}
\usepackage{xcolor}
\usepackage{hyperref}
\usepackage[nocompress]{cite}
\hypersetup{
    colorlinks,
    linkcolor=blue,
    citecolor=blue,
    urlcolor=blue
}

\urldef{\mailsa}\path|{qian.wang, ke.chen}@manchester.ac.uk|
\newcommand{\keywords}[1]{\par\addvspace\baselineskip
\noindent\keywordname\enspace\ignorespaces#1}

\begin{document}

\mainmatter  % start of an individual contribution

\title{Alternative Semantic Representations for Zero-Shot Human Action Recognition}

\titlerunning{Alternative Semantic Representations for Zero-Shot Action Recognition}

\author{Qian Wang \textsuperscript{(\Letter)} %
%\thanks{footnote.}%
\and Ke Chen}
\authorrunning{Q. Wang and K. Chen}

\institute{School of Computer Science, The University of Manchester,\\
Manchester, M13 9PL, UK\\
\mailsa\\
%\url{http://www.springer.com/lncs}
}


\toctitle{Lecture Notes in Computer Science}
\tocauthor{Authors' Instructions}
\maketitle

\begin{abstract}
    A proper semantic representation for encoding side information is key to the success of zero-shot learning. In this paper, we explore two alternative semantic representations especially for zero-shot human action recognition:  textual descriptions of human actions and deep features extracted from still images relevant to human actions. Such side information are accessible on Web with little cost, which paves a new way in gaining side information for large-scale zero-shot human action recognition. We investigate different encoding methods to generate semantic representations for human actions from such side information. Based on our zero-shot visual recognition method, we conducted experiments on UCF101 and HMDB51 to evaluate two proposed semantic representations . The results suggest that our proposed text- and image-based semantic representations outperform traditional attributes and word vectors considerably for zero-shot human action recognition. In particular, the image-based semantic representations yield the favourable performance even though the representation is extracted from a small number of images per class.

\end{abstract}

\section{Introduction}
\begin{figure}
\includegraphics[width=1\linewidth]{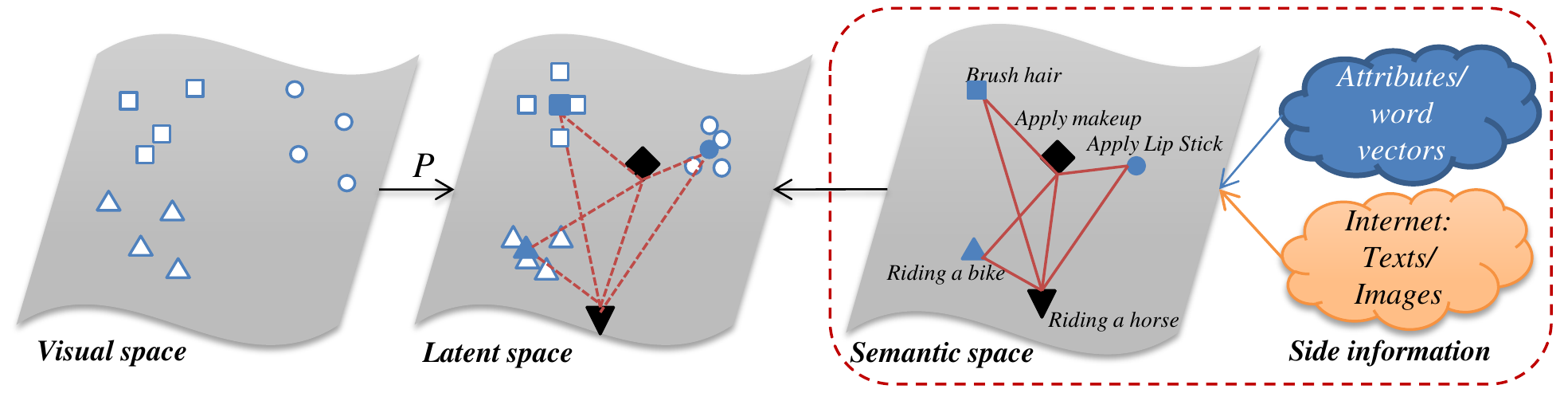}
\caption{A schematic diagram of zero-shot learning framework. The work in this paper is highlighted in the dashed box. Human action classes are denoted by coloured markers (blue and black for training and unseen classes respectively) with different shapes. The training data are used to learn the mapping $P$ and training class embedding (blue filled markers in the latent space), then the unseen class embedding (black filled markers in the latent space) is achieved by preserving the semantic distances (red lines). See Section \ref{sect_bidilel} for more details of our ZSL method.}
\label{fig_scheme}
\end{figure}

Zero-Shot Learning (ZSL) aims to recognize instances from new classes which are not seen in the training data. It is a promising alternative to the traditional supervised learning which requires labour-intensive annotation work on all the classes involved. As shown in Figure \ref{fig_scheme}, in ZSL, the knowledge learned from training data is transferred to recognise unseen classes through the side information which can usually be acquired with less effort. Although most existing works in ZSL focus on the development of novel recognition models, the side information for knowledge transfer plays an equally important role in the success of ZSL. The most popular side information used in ZSL literature are attributes and word vectors. Although they have been widely used in ZSL \cite{lampert2009learning,liu2011recognizing,inoue2016adaptation,wang2016zero,xian2017zero}, both of them have obvious drawbacks as well, especially for zero-shot human action recognition in video data.

The definition and annotation of attributes for human actions (e.g., the attributes defined for UCF101 \cite{jiang2014thumos} include ``bodyparts-visible: face, fullbody, onehand", ``body-motion: flipping, walking, diving, bending", etc.) are subjective and labour-intensive. When a large number of human actions are involved, more attributes are needed to distinguish one human action from the other. As a result, attributes based semantic representations are inappropriate for large scale zero-shot human action recognition.
On the other hand, as stated in \cite{alexiou2016exploring}, using a word vector of the class label to represent a human action is far from adequate to illustrate the rich appearance variations. In addition, the word vectors are learned from textual corpus, thus suffering from the catastrophic semantic gap problem (i.e., the difference of information conveyed by visual media and texts).

To address the limitations of existing semantic representations for ZSL, we attempt to explore alternative side information towards enhanced zero-shot human action recognition. The essentials of side information for ZSL are twofold. Firstly, it should be achievable for a large number of human actions without much effort. More importantly, the side information should be able to capture the visually discriminative semantics thus benefiting the ZSL by easily bridging the semantic gap. To this end, we employ action relevant images as the side information resources to extract the semantics of human actions. With the aid of search engines, it is effortless to collect a set of action relevant images by using the action name as the key words. Although still images lack of temporal information in human actions, they provide abundant visually discriminative information which can be exploited to extract high-level semantic representations for human actions. On the other hand, we aim to enhance the word vectors by collecting and encoding textual descriptions of human actions. We believe that the contextual information in the action relevant texts (e.g., description articles of human actions from the web) will remove the ambiguity of the semantics in the original action word vectors which are based solely on the action labels.

To summarise, the contributions of this paper include:
\begin{itemize}
\item We propose and implement the idea of using textual descriptions to enhance the word vector representations of human actions in ZSL.
\item We propose and implement the idea of using action related still images to represent semantics for video based human actions in ZSL.
\item Experiments are conducted to evaluate the effectiveness of the proposed semantic representations in zero-shot human action recognition, and significant performance improvement has been achieved.
\end{itemize}
\section{Related Work}

The semantic representation is key for the success of ZSL. Recently, attempts have been made to explore more effective semantic representations for objects/ actions towards improved ZSL performance. In this section, we will review the prevailing semantic representations used in ZSL (Table \ref{table_survey}), including a variety of extensions of attributes and word vectors, as well as many other less popular approaches proposed in literature.

\begin{table*}[t]
{\normalsize
\centering
\caption[]{A survey on semantic representations in ZSL}
\label{table_survey}
\newsavebox{\tablebox}
\begin{lrbox}{\tablebox}
\begin{tabular}{@{}ll}\toprule
  \textbf{Authors and Year} & \textbf{Semantic Representation} \\  \midrule
  Lampert \textit{et al}. (2009)\cite{lampert2009learning}  & Attributes, annotated manually           \\
  Sharmanska \textit{et al}. (2012)\cite{sharmanska2012augmented} & Attributes, enhanced by learning from visual data \\
  Liu \textit{et al}. (2011) \cite{liu2011recognizing} & Attributes, enhanced by learning from visual data\\
  Qin \textit{et al}. (2016) \cite{qin2016beyond} & Attributes, enhanced by learning from visual data \\
  Fu \textit{et al}. (2014) \cite{fu2014learning} & Attributes, enhanced by learning from visual data \\ \midrule
  Inoue \textit{et al}. (2016) \cite{inoue2016adaptation} & Word vector, enhanced by a weighted combination\\
  &\hspace{4ex} of related word (from \textit{WordNet}) vectors.\\
  Alexiou \textit{et al}. (2016) \cite{alexiou2016exploring} & Word vector, enhanced by the synonyms of labels \\
                                                    & \hspace{4ex}(from Internet dictionaries) \\
  Mukherjee \textit{et al}. (2016) \cite{mukherjee2016gaussian} & Word Gaussian distribution \\
  Sandouk \textit{et al}. (2016) \cite{sandouk2016multi} & Word vector, enhanced by contexts (from tags) \\ \midrule
  Elhoseiny \textit{et al}. (2013) \cite{elhoseiny2013write} & Tf-idf, based on \textit{Wikipedia} articles \\
  Akata \textit{et al}. (2016) \cite{akata2016multi} & BOW, based on \textit{Wikipedia} articles\\
  Rohrbach \textit{et al}. (2010) \cite{rohrbach2010helps} & \textit{WordNet} path length, based on \textit{WordNet} ontology\\
                                                  & \hspace{4ex} Hit-counts, based on web search results\\
  Chuang \textit{et al}. (2015) \cite{chuang2015exploring} & \textit{WordNet} path length, based on \textit{WordNet} ontology\\
  
\bottomrule
\end{tabular}
\end{lrbox}
\scalebox{0.8}{\usebox{\tablebox}}
}
\end{table*}

Attributes based semantic representations were firstly proposed for ZSL in \cite{lampert2009learning}, thereafter, attributes have been employed for ZSL in many works \cite{xian2017zero,zhang2016zero,chao2016empirical,wang2016zero}. A set of binary attributes need to be manually defined to represent the semantic properties of objects. As a result, each object class can be represented by a binary attribute vector in which the value of one and zero indicates the presence and absence of each attribute respectively. Since the attributes are shared by seen and unseen classes, the knowledge transfer is enabled. However, as mentioned above, the definition of attributes require experts with domain knowledge to discriminate different classes, and the attribute annotation for a large number of classes could be subjective and labour-intensive.

Alternatively, attributes can be mined automatically from visual features by discriminative mid-level feature learning \cite{farhadi2009describing,liu2011recognizing,sharmanska2012augmented,qin2016beyond,fu2014learning}, but their semantic meanings are unknown, thus inappropriate for direct use in ZSL. To enhance the attributes' discriminative power and semantic meaningfulness, the manually defined attributes and the ones automatically learned from training data are usually combined. However, the data-driven attributes are usually dataset specific and probably fail on a different dataset.

The other kind of prevailing side information used in ZSL is derived from text resources. One of the most popular semantic representations is word vector (e.g., the ones generated by the \textit{word2vec} tool \cite{mikolov2013distributed}) due to its convenience and effectiveness. A class label can be easily represented with the vector representation of the corresponding word or phrase. However, word vectors are deficient to discriminate different classes from the visual perspective due to the semantic gap, i.e., the gap between visual and semantic information. As a result, word vectors are usually outperformed by attributes in ZSL.

To alleviate the semantic gap problem, some attempts have been made to enhance the word vectors \cite{inoue2016adaptation,alexiou2016exploring,mukherjee2016gaussian,sandouk2016multi}.  Inoue \textit{et al}. \cite{inoue2016adaptation} aim to adapt the original word vectors to make two visually similar concepts close to each other in the adapted word vector space by representing a concept with a weighted sum of its original word vector and its hypernym (based on \textit{WordNet}) word vectors. And the weights are learned from visual resources. Alexiou \textit{et al}. \cite{alexiou2016exploring} enrich the word vector representation by mining and considering synonyms of the action class labels from multiple \textit{Internet dictionaries}.  Mukherjee \textit{et al}. \cite{mukherjee2016gaussian} use \textit{Gaussian distribution} instead of a single word vector to model the class labels so that the intra-class variability can be expressed properly in the semantic representations. To address the issue of polysemy, Sandouk \textit{et al}. \cite{sandouk2016multi} learn a specific vector representation for a word together with its context. That is to say, the same word could have different vector representations when it is in different contexts. Inspired by these works, our work further investigates the possible side information and enabling techniques to enhance the word vectors for ZSL.

Other than attributes and word vectors, other side information has also been investigated for knowledge transfer in ZSL, only if they are able to model the relationships among different classes and relatively easy to obtain. For example, \textit{WordNet} path length is used to measure the semantic correlations between two concepts in \cite{chuang2015exploring,rohrbach2010helps}. The Internet together with search engines provides a natural opportunity to get side information to measure between-class semantic relationships based on hit-count on search results \cite{rohrbach2010helps}. Textual descriptions of a class rather than the single class name are employed to represent a class in \cite{elhoseiny2013write,akata2016multi}. Concept related textual descriptions (e.g., \textit{Wikipedia} page) can be readily obtained from the Internet and then processed with techniques in natural language processing (NLP). Considering our focus on zero-shot human action recognition based on video data, images from the Internet can be alternative side information to texts which have been a typical choice for zero-shot image classification.

\section{Method}
\label{method}
\begin{figure}
\includegraphics[width=1\linewidth]{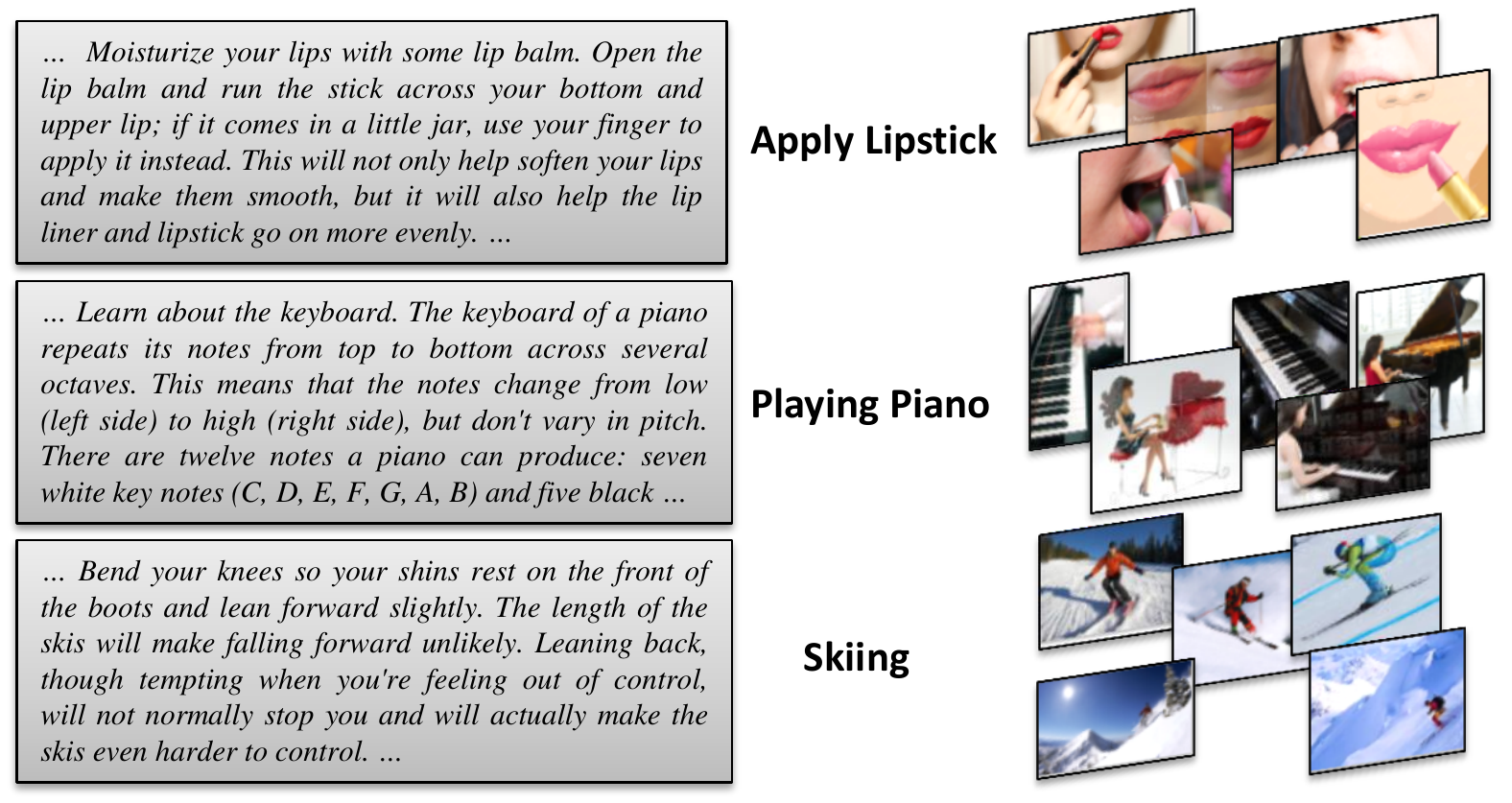}
\caption{Examples of collected description texts and images of three human actions from UCF101 (i.e., ``Apply Lipstick", ``Playing Piano" and ``Skiing").}
\label{fig_examples}
\end{figure}

In this section, we propose our methods of generating semantic representations for zero-shot human action recognition from text and image resources respectively. Firstly, we use search engines to collect action relevant texts and images as the side information. Some typical examples are shown in Fig.\ref{fig_examples}. Once the side information are collected, we use different encoding approaches to generate the semantic representations for human actions.
\subsection{Text-based Semantic Representation}
\label{sect_texts}
\subsubsection{Texts Collection}
Motivated by the fact a class label is insufficient to depict the complex concepts in the human action, we try to collect textual descriptions from the web to represent each human action. Textual descriptions of human actions can be derived from \textit{WikiHow}, a website teaching people ``how to do anything". Inevitably, the description texts for some actions (e.g., ``pick", ``sit") are not available from \textit{WikiHow}, for which we turn to alternative sources including \textit{Wikipedia} and \textit{Online dictionary}. 

\subsubsection{Pre-Processing}
Once the textual descriptions for all the human actions are collected, we end up with a document for each human action class. We use natural language processing techniques to pre-process the unstructured textual data before encoding them into semantic representations. In the first step, we tokenize the documents to get all the words appearing in the documents. After removing the stop words (i.e., the words carrying little semantic meanings such as ``is", ``you", ``of"), we have a dictionary containing $d$ words.

\subsubsection{Term-Document Matrix (TD)}
Given the documents and the dictionary containing all the terms/words in the documents, a term-document matrix $M$ is constructed to represent the term frequency in all documents. $M_{ij}$  denotes the frequency of term $i$ in document $j$, where $i=1,2,...,d$ and $j=1,2,...,C$, $C$ is the number of documents, i.e., the number of human actions in a specific dataset. Thus the column vectors in $M$ can be used to represent the semantic representations of human actions. We denote this approach as \textbf{TD} in the following sections.
\subsubsection{Average Word Vector (AWV)}
We aim to enhance the word vectors by incorporating the collected textual information. Taking advantage of the compositional property of word vectors, we can represent a document with the average of all the included word vectors.
\begin{equation}
\label{eq_awv}
AWV(j) = \frac{1}{n_j}\sum_{i=1}^{n_j} v_i
\end{equation}
where $n_j$ is the number of terms in the $j$-th document, $v_i\in \mathbb{R}^D$ denotes the word vector of the $i$-th term in the document, and $D$ is the dimensionality of word vectors.
\subsubsection{Fisher Word Vector (FWV)}
\label{fwv}
In contrast to AWV using the mean of all word vectors to represent a document, FWV aims to model the distribution of word vectors in a document. Fisher Vector represents a document (i.e., a set of words) by the gradient of log likelihood with respect to the parameters of a pre-learned probabilistic model (i.e., Gaussian Mixture Model) \cite{perronnin2010improving,vedaldi08vlfeat}. A Gaussian Mixture Model (GMM) is used to fit the distribution of the word vectors involved in all documents, where the parameters $\Theta=\{\mu_k, \Sigma_k, \pi_k\}, k=1,...,K$. Let $V^j=\{v_1,...,v_{n_j}\}$ be a set of word vectors from the $j$-th human action description document. Then the Fisher Vector of $j$-th document can be denoted by:
\begin{equation}
\label{eq_fwv}
FWV(j)=[\mathcal{G}_{\mu,1}^{V^j},...,\mathcal{G}_{\mu,K}^{V^j},\mathcal{G}_{\sigma,1}^{V^j},...,\mathcal{G}_{\sigma,K}^{V^j}],
\end{equation}
where
\begin{equation}
\label{eq_mu}
\mathcal{G}_{\mu,k}^{V^j} = \frac{1}{\sqrt{\pi_k}}\sum_{v_i\in {V^j}} \gamma_{k,i} (\frac{v_i-\mu_k}{\sigma_k}),
\end{equation} 
\begin{equation}
\label{eq_sigma}
\mathcal{G}_{\sigma,k}^{V^j} = \frac{1}{\sqrt{2\pi_k}}\sum_{v_i\in {V^j}} \gamma_{k,i} (\frac{(v_i-\mu_k)^2}{\sigma_k^2}-1),
\end{equation}
\begin{equation}
\label{eq_gamma}
\gamma_{k,i}=\frac{exp[-\frac{1}{2}(v_i-\mu_k)^T\Sigma_k^{-1}(v_i-\mu_k)]}{\Sigma_{t=1}^K exp[-\frac{1}{2}(v_i-\mu_t)^T\Sigma_k^{-1}(v_i-\mu_t)]}.
\end{equation}
The dimension of the Fisher Vector is $2DK$, where $D$ and $K$ are the dimensionality of word vectors and the number of components in the GMM respectively.
\subsection{Image-based Semantic Representation}
\label{sect_images}
Human actions are difficult to describe with texts due to the complexity and intra-class variations. Although they lack temporal information, still images can provide abundant information for the understanding of human actions. Compared to the video examples, still images are much easier to collect, annotate and store. Thus we hold the view that still images are a proper kind of side information which can benefit modelling human action relationships with little effort.
\subsubsection{Image Collection}
Given a human action, we use the label as the key word and search relevant images with search engines. For most human actions we can get a collection of images each of which gives a view of the action. However, for some action names which could have multiple meanings, the additional explaining key words are needed to get reasonable searching results. For example, we use ``salsa spin + dancing" and ``playing + hula hoop" for the actions ``salsa spin" and ``hula hoop" respectively. For each human action, we get different numbers of relevant images after removing the ones of poor quality (e.g., irrelevant ones and the ones smaller than 10Kb) from the returned results. The image collection and filtering can be processed automatically without many human interventions \footnote{The image scraper tool is available: http://staff.cs.manchester.ac.uk/~kechen/ASRHAR/}.
\subsubsection{Feature Extraction}
We aim to extract useful information from a set of images to represent a human action. Recently, deep convolution neural networks have been used to extract image features carrying high-level conceptual information. By feeding the images into a pre-trained CNN model, the deep image features can be obtained easily. Then each human action is represented with a set of image feature vectors $F^j = \{f_1,...,f_{n_j}\}$. In the next two sections, we use two approaches to encode the set of image features into the action-level semantic representation.
\subsubsection{Average Feature Vector (AFV)}
Similar to Eq.(\ref{eq_awv}), we can use the average of multiple image features as the human action semantic representation.
\begin{equation}
\label{eq_afv}
AFV(j) = \frac{1}{n_j} \sum_{i=1}^{n_j} f_i
\end{equation}
\subsubsection{Fisher Feature Vector (FFV)}
Similar to the processing applied on word vectors in Section \ref{fwv}, we use Fisher Vector to encode a set of image feature vectors relevant to a specific human action.
\begin{equation}
\label{eq_ffv}
FFV(j)=[\mathcal{G}_{\mu,1}^{F^j},...,\mathcal{G}_{\mu,K}^{F^j},\mathcal{G}_{\sigma,1}^{F^j},...,\mathcal{G}_{\sigma,K}^{F^j}],
\end{equation}
where $\mathcal{G}_{\mu,i}^{F^j}$ and $\mathcal{G}_{\sigma,i}^{F^j}$ can be calculated in the same way as Eq.(\ref{eq_mu}-\ref{eq_gamma}).

\section{Experimental Settings}
\subsection{Dataset}
We use two human action datasets to evaluate the proposed approaches for zero-shot recognition, i.e., UCF101 \cite{soomro2012ucf101} and HMDB51 \cite{kuehne2011hmdb}. \textbf{UCF101} is a human action recognition dataset collected from YouTube. There are 13,320 real action video clips falling into 101 action categories. In our experiments, we use 5 randomly generated 51/50 (seen/unseen) class-wise data splits. \textbf{HMDB51} contains 6,766 video clips from 51 human action classes. Similarly, we use 5 randomly generated 26/25 splits in all experiments.
\subsection{Zero-Shot Recognition Method}
\label{sect_bidilel}
We employ our recently developed ZSL method, bidirectional latent embedding learning (BiDiLEL) \cite{wang2016zero}, as a test bed in our experiments \footnote{Like attributes and word vectors, our proposed semantic representations may be directly deployed in all the existing zero-shot human action recognition methods.}. To make the paper self-contained, we will briefly describe the main idea of BiDiLEL in this section.

The method employs a two-stage latent embedding algorithm to learn a latent space in which the semantic gap is bridged and zero-shot recognition can be done (see Fig.\ref{fig_scheme}). In bottom-up stage, we learn a projection matrix $P$ by supervised locality preserving projection (SLPP) \cite{cheng2005supervised}, such that the examples close to each other in the original visual space will still be close in the latent space. By exploiting the local structures and labelling information in the training data, the learned latent space preserves the data distribution and is more discriminative. The properties are expected to generalise well for test examples from unseen classes.

In the top-down embedding, the latent embedding of each seen class can be calculated by averaging the projections of all the training examples from the class and then serve as landmarks guiding the learning of latent embedding of unseen classes. We use the landmarks based Sammon mapping (LSM) \cite{wang2016zero} which aims to preserve the inter-class semantic distances (measured in the semantic space). As a result, the semantic distances between seen and unseen classes as well as between any pair of unseen classes will be preserved in the latent space.

Once the latent embedding of both seen and unseen classes are obtained, we can do the zero-shot learning in the latent space using the nearest neighbour method. Specifically, given a test example, we use projection matrix $P$ to map it into the latent space, where its distances to all the class embedding can be calculated, and it will be assigned to the closest class label. For more details, we refer the readers to \cite{wang2016zero}.

\subsection{Video Representation}
C3D was proposed in \cite{tran2015learning} for human action recognition. It utilizes 3D ConvNets to learn spatio-temporal features for video streams. According to \cite{wang2016zero}, the C3D video representation outperforms its counterparts in zero-shot human action recognition. We use the model pre-trained on Sports-1M dataset and follow the setting in \cite{tran2015learning,wang2016zero} to extract spatio-temporal deep features (i.e., the 4096-dimensional ``fc6" activations of the deep neural network) from 16-frame segments. Finally, the visual representation of a video stream is calculated by averaging the features of all the segments from the video.

\subsection{Evaluation}
\label{sect_eval}

In most existing ZSL works, the evaluations are based on the assumption that test examples are only from unseen classes, which is often referred as to conventional zero-short learning (cZSL). In practice, however, the test examples can be from either training classes or unseen classes. To evaluate ZSL methods in a more practical scenario, the problem of generalised ZSL has been formulated and investigated in \cite{chao2016empirical,xian2017zero}. In gZSL, given a test example, the label search space consists of both seen and unseen classes. In our experiments, we follow the protocols in \cite{xian2017zero} and report both conventional and generalised ZSL (cZSL and gZSL) results using per-class accuracy. In the generalised ZSL scenarios, except the examples from test classes, we also reserve 20\% examples from each training class for testing and the rest 80\% examples from each training class for training. 

Concretely, we report the recognition accuracy of test examples from unseen classes by setting the search space in the unseen label set $\mathcal{U}$ for the cZSL; the accuracy is denoted by $A_{\mathcal{U}\rightarrow \mathcal{U}}$. For gZSL, we set the search space in the whole label set $\mathcal{T}=\mathcal{S}\cup \mathcal{U}$ and report three types of per-class accuracies, i.e., the recognition accuracy of test examples from unseen classes $A_{\mathcal{U}\rightarrow \mathcal{T}}$, the recognition accuracy of test examples from seen classes $A_{\mathcal{S}\rightarrow \mathcal{T}}$ and the harmonic mean,
\begin{equation}
H = 2*A_{\mathcal{U}\rightarrow \mathcal{T}}*A_{\mathcal{S}\rightarrow \mathcal{T}}/(A_{\mathcal{U}\rightarrow \mathcal{T}}+A_{\mathcal{S}\rightarrow \mathcal{T}}).
\end{equation}

The ZSL method employed in our experiments works in the inductive setting (i.e., the test example is processed individually), but can be extended to the transductive setting (i.e., all the test examples are assumed to be available as a collection when doing the recognition) easily by using the structured prediction method\cite{wang2016zero,zhang2016zero}. The method of structure prediction uses Kmeans to group all the test examples into clusters (the number of clusters is set to be the number of unseen classes) and find a one-to-one map from the clusters to unseen classes. In our experiments, we will report the results of cZSL in both inductive and transductive settings.
 
\section{Experimental Results}
In this section, we present the designed experiments and the results to evaluate the effectiveness of proposed semantic representations \footnote{The scripts and data used in our experiments can be available on our project page: http://staff.cs.manchester.ac.uk/~kechen/ASRHAR/}.

\subsection{Text-based Representation}
\label{sect_texts_results}
We conduct experiments of zero-shot human action recognition by utilising the proposed text-based semantic representations in Section \ref{sect_texts}, i.e., TD, AWV and FWV. We use the 300-dimensional word vectors pre-trained with \textit{word2vec} on Google News dataset (about 100 billion words) \footnote{https://code.google.com/p/word2vec/}. For FWV, we set the value of $K$ in Eq.(\ref{eq_fwv}) to be $\{1,2,3,4,5\}$. The experiments aim to investigate how different text-based semantic representations perform in zero-shot human action recognition. In our experiments, we follow the protocols in \cite{wang2016zero} using class-wise cross validation to find the optimal values of hyper-parameters. According to the performance on the validation data, cosine distances are employed to calculate the semantic distances for FWV, and Euclidean distances are employed for AWV.

We report the results of conventional ZSL in both inductive and transductive settings in Table \ref{table_texts}. With only the textual description sources, the simple encoding method TD can achieve the accuracy of 19.54\% and 15.26\% respectively on UCF101 and HMDB51, which indicates the textual descriptions collected by search engines are useful for modelling the inter-class relationships. By incorporating the pre-trained word vectors, AWV improves the accuracy to 24.38\% and 21.80\% respectively on UCF101 and HMDB51. On the other hand, by comparing FWV with different $K$ values, we know that $K=1$ gives the best results with an accuracy of 23.76\% on UCF101 and 19.57\% on HMDB51; however, it is still outperformed by AWV on both datasets regardless of inductive or transductive settings. To conclude, AWV performs the best among different text-based semantic representations.

\begin{table*}[t]
{\normalsize
\centering
\caption[]{Results of different text-based semantic representations (mean$\pm$standard error of recognition accuracy \%) on UCF101 and HMDB51 datasets. (Sem.Rep.--Semantic Representation, Att--Attributes, WV--Word vector)
}
\label{table_texts}

\begin{lrbox}{\tablebox}
\begin{tabular}{@{}lccccc}\toprule
  \multirow{2}{*}{\textbf{Sem. Rep.}} & \multicolumn{2}{c}{\textbf{UCF101 (51/50)}} & \multicolumn{2}{c}{\textbf{HMDB51 (26/25)} }\\
  \cmidrule(l){2-5}
  & \textbf{Inductive} & \textbf{Transductive} & \textbf{Inductive} & \textbf{Transductive}  \\ \midrule
Random  & $2.00$		   & $2.00$          & $4.00$          & $4.00$ \\
Att		& $21.54\pm 0.72$  & $\mathbf{32.00\pm 2.30}$ & -			   & - \\
WV		& $19.42\pm 0.69$  & $22.05\pm 1.74$ & $21.53\pm 1.75$ & $24.14\pm 3.43$ \\
TD	     & $19.54\pm 0.75$ & $24.29\pm 0.65$ & $15.26\pm 0.57$ & $15.33\pm 1.72$ \\
AWV 	 & $\mathbf{24.38\pm 1.00}$ & $30.60\pm 2.67$ & $\mathbf{21.80\pm 0.87}$ & $\mathbf{26.13\pm 1.29}$ \\
FWV(K=1) & $23.76\pm 0.72$ & $28.54\pm 0.70$ & $19.57\pm 1.21$ & $20.41\pm 1.74$ \\
FWV(K=2) & $23.61\pm 1.08$ & $28.64\pm 1.45$ & $18.80\pm 1.22$ & $20.01\pm 1.74$ \\
FWV(K=3) & $22.21\pm 0.96$ & $24.33\pm 2.34$ & $17.35\pm 1.93$ & $21.37\pm 3.16$ \\
FWV(K=4) & $22.11\pm 0.62$ & $28.76\pm 1.03$ & $17.07\pm 1.41$ & $18.80\pm 2.95$ \\
FWV(K=5) & $21.50\pm 0.67$ & $27.56\pm 2.43$ & $16.95\pm 1.19$ & $17.20\pm 1.92$ \\
\bottomrule
\end{tabular}
\end{lrbox}
\scalebox{0.8}{\usebox{\tablebox}}
}
\end{table*}

\subsection{Image-based Representation}
In our experiments, we collect variant numbers of relevant images for different human actions. The average number of relevant images per class is around 200 and 100 for UCF101 and HMDB51 respectively. To extract the image features, we use the GoogLeNet \cite{szegedy2015going} model pre-trained on ImageNet dataset \footnote{http://www.vlfeat.org/matconvnet/}. The activations of top fully connected layer of GoogLeNet of 1024 dimensions are used as the deep image features. We evaluate the image-based semantic representations encoded with different approaches described in Section \ref{sect_images}, i.e., AFV and FFV. Again, we set the values of $K$ in Eq.(\ref{eq_ffv}) to be $\{1,2,3,4,5\}$. We employ the same experiment protocols as those used in the previous experiments (Section \ref{sect_texts_results}). According to the performance on the validation data, cosine distances are employed to model the semantic distances for FFV, and Euclidean distances are employed for AFV.

The experimental results are shown in Table \ref{table_images}. Apparently, $K=1$ again gives the best performance of FFV, achieving 40.12\% and 25.82\% respectively on UCF101 and HMDB51 in the inductive setting, 50.67\% and 31.51\% respectively on UCF101 and HMDB51 in the transductive setting. Different from the text-based semantic representations, image-based semantic representations FFV encoded by Fisher Vector outperforms the AFV on both datasets.

\begin{table*}[t]
{\normalsize
\centering
\caption[]{Results of different image-based semantic representations (mean$\pm$standard error of recognition accuracy \%) on UCF101 and HMDB51 datasets.
}
\label{table_images}

\begin{lrbox}{\tablebox}
\begin{tabular}{@{}lccccc}\toprule
  \multirow{2}{*}{\textbf{Sem. Rep.}} & \multicolumn{2}{c}{\textbf{UCF101 (51/50)}} & \multicolumn{2}{c}{\textbf{HMDB51 (26/25)} }\\
  \cmidrule(l){2-5}
  & \textbf{Inductive} & \textbf{Transductive} & \textbf{Inductive} & \textbf{Transductive}  \\ \midrule
Random  & $2.00$		   & $2.00$          & $4.00$          & $4.00$ \\
AFV 	 & $37.24\pm 0.89$ & $50.48\pm 1.35$ & $25.55\pm 1.66$ & $30.77\pm 3.23$ \\
FFV(K=1) & $\mathbf{40.12\pm 1.30}$ & $\mathbf{50.67\pm 2.45}$ & $\mathbf{25.82\pm 1.19}$ & $\mathbf{31.51\pm 1.67}$ \\
FFV(K=2) & $38.01\pm 1.58$ & $49.60\pm 1.82$ & $25.50\pm 0.95$ & $28.98\pm 1.94$ \\
FFV(K=3) & $36.52\pm 1.38$ & $45.48\pm 0.73$ & $24.27\pm 1.10$ & $26.95\pm 3.38$ \\
FFV(K=4) & $35.31\pm 1.17$ & $44.76\pm 2.40$ & $23.22\pm 1.25$ & $25.26\pm 2.32$ \\
FFV(K=5) & $34.98\pm 0.68$ & $45.08\pm 1.82$ & $23.09\pm 1.12$ & $23.93\pm 2.06$ \\
\bottomrule
\end{tabular}
\end{lrbox}
\scalebox{0.8}{\usebox{\tablebox}}
}
\end{table*}

\subsection{Comparison with Other Semantic Representations}
In this experiment, we compare the proposed semantic representations with other popular ones. From Table \ref{table_texts} and \ref{table_images}, we know that AWV and FFV(K=1) perform the best among the text- and image-based semantic representations respectively. So we consider AWV and FFV(K=1) as the representatives of the proposed text- and image-based semantic representations. As described in Section \ref{sect_eval}, we conduct the experiments in both conventional and generalised ZSL scenarios in our experiments.

We present the experimental results in Table \ref{table_comparison}. Clearly, the proposed two semantic representations (i.e., AWV and FFV(K=1)) outperform word vectors and attributes consistently in terms of the conventional ZSL evaluation metric. On UCF101, the use of textual information enhances the word vectors based solely on the action labels by lifting the accuracy from 19.42\% to 24.38\%, even higher than that of labour-intensive attributes (21.54\%). The image-based semantic representation FFV encoded with Fisher Vector gives the best accuracy of 40.12\%, significantly higher than its counterparts. This is attributed to the narrower semantic gap between video representation space and image-based semantic space. The still images contain abundant visually discriminative information which can be further encoded into high-level semantic representations of human actions. On HMDB51, the same conclusions can be drawn. It is noteworthy that AWV is only slightly better than WV for HMDB51 dataset. The reason might be the existence of actions which are difficult to describe with texts in this dataset, such as, ``sit", ``talk", ``turn", ``stand", ``pick", ``catch", and etc.

Regarding the generalised ZSL scenario, the proposed AWV and FFV perform better on the test examples from unseen classes (with 5.32\% and 16.55\% respectively on UCF101, 2.99\% and 5.91\% respectively on HMDB51), outperforming the attributes and word vectors. We also notice that FFV does not perform the best on test examples from seen classes (i.e., $A_{\mathcal{S} \rightarrow \mathcal{T}}$), although it is significantly better than others in terms of harmonic mean (H). This is reasonable and practically preferable with the trade-off between recognition accuracy of examples from seen and unseen classes.

\begin{table*}[t]
{\normalsize
\centering
\caption[]{A comparison of different semantic representations  on UCF101 and HMDB51 datasets (mean$\pm$standard error)\%.
}
\label{table_comparison}

\begin{lrbox}{\tablebox}
\begin{tabular}{@{}lc|c|ccc}\toprule
  \multirow{2}{*}{\textbf{Dataset}} & \multirow{2}{*}{\textbf{Sem. Rep.}} & \multicolumn{1}{c|}{\textbf{cZSL}} & \multicolumn{3}{c}{\textbf{gZSL} }\\
  \cmidrule(l){3-6}
  & & \textbf{$A_{\mathcal{U} \rightarrow \mathcal{U}}$} & \textbf{$A_{\mathcal{U} \rightarrow \mathcal{T}}$} & \textbf{$A_{\mathcal{S} \rightarrow \mathcal{T}}$} & {H}  \\ \midrule
  
\multirow{5}{*}{\textbf{UCF101}}& Random  & $2.00$		   & $1.00$          & $1.00$          & $1.00$ \\
& WV 	 & $19.42\pm 0.69$ & $4.54\pm 0.64$ & $84.79\pm 0.91$ & $8.59\pm 1.17$ \\
& Att & $21.54\pm 0.72$ & $2.48\pm 0.62$ & $86.39\pm 1.37$ & $4.78\pm 1.18$ \\
& AWV & $24.38\pm 1.00$ & $5.32\pm 1.53$ & $\mathbf{86.43\pm 1.06}$ & $9.85\pm 2.66$ \\
& FFV & $\mathbf{40.12\pm 1.30}$ & $\mathbf{16.55\pm 1.30}$ & $82.38\pm 1.17$ & $\mathbf{27.49\pm 1.86}$ \\
%& AWV+FFV & & & & \\
\midrule

\multirow{4}{*}{\textbf{HMDB51}}& Random  & $4.00$		   & $2.00$          & $2.00$          & $2.00$ \\
& WV 	 & $21.53\pm 1.75$ & $2.64\pm 0.33$ & $58.70\pm 1.40$ & $5.05\pm 0.61$ \\
%& Att & $\mathbf{40.12\pm 1.30}$ & $\mathbf{50.67\pm 2.45}$ & $\mathbf{25.68\pm 1.07}$ & $\mathbf{33.18\pm 2.09}$ \\
& AWV & $21.80\pm 0.87$ & $2.99\pm 0.35$ & $\mathbf{62.00\pm 2.57}$ & $5.69\pm 0.64$ \\
& FFV & $\mathbf{25.68\pm 1.07}$ & $\mathbf{5.91\pm 0.90}$ & $58.57\pm 1.50$ & $\mathbf{10.65\pm 1.48}$ \\
%& AWV+FFV & & & & \\
\bottomrule
\end{tabular}
\end{lrbox}
\scalebox{0.8}{\usebox{\tablebox}}
}
\end{table*}

\subsection{How Many Images Are Enough?}
\label{sect_noImgs}
In the previous experiments, we use all the collected images to encode the image-based semantic representations. In this experiment, we investigate how the number of images affects the encoded semantic representations. We use AFV and FFV(K=1) as the encoding methods and generate the semantic representations for each human action with the number of relevant images to be 5, 10, 20, 30, 40, 50, 60, 70, 80, 90 and 100 respectively (For the case when the total number of collected images for one human action is less than the expected number, we simply use all the collected images of that action in the experiment). The experiments are conducted on two human action datasets in conventional ZSL scenario under both inductive and transductive settings.

\begin{figure}
\includegraphics[width=1\linewidth]{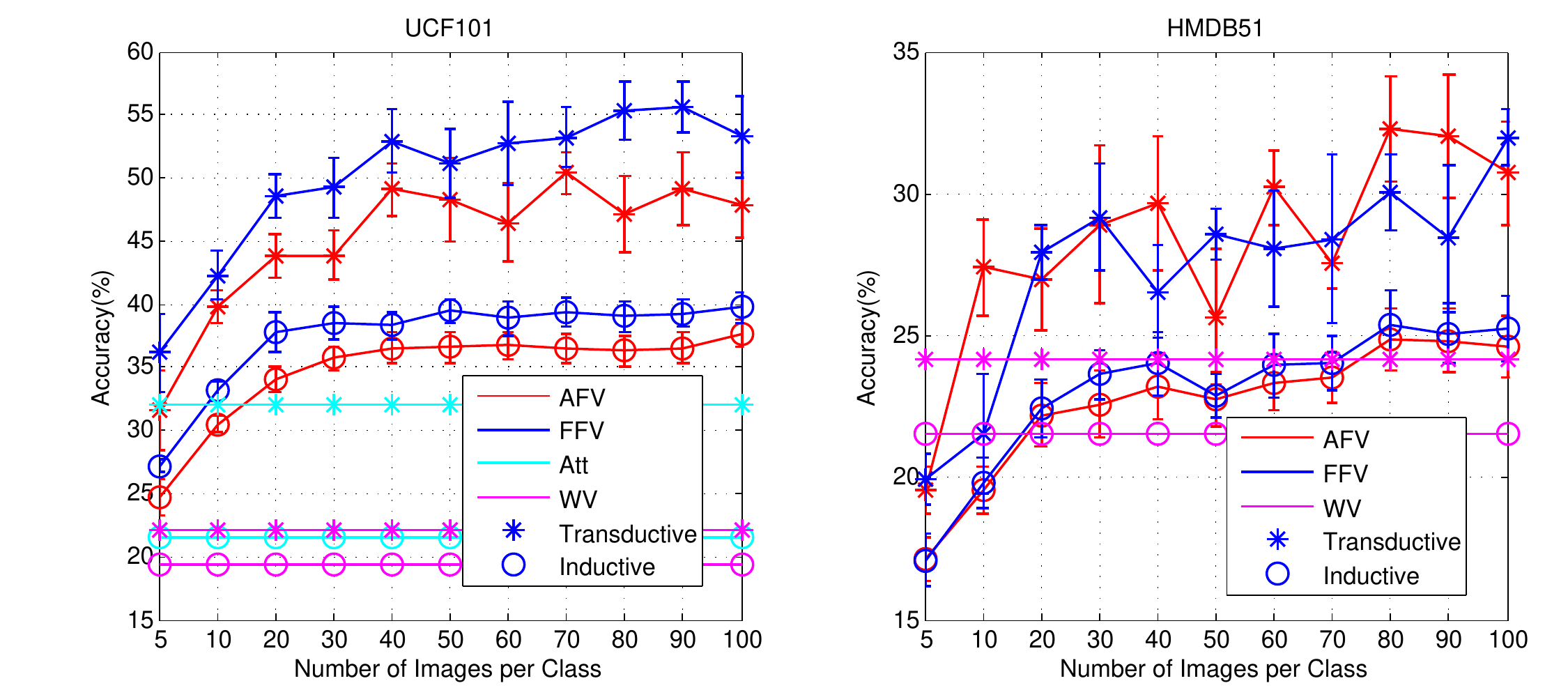}
\caption{Effects of number of images on the performance of AFV and FFV(K=1).}
\label{fig_noImgs}
\end{figure}

The performances of two types of image-based semantic representations with different numbers of images are shown in Fig.\ref{fig_noImgs}. For a direct comparison, we display the baseline performance of attributes and word vectors in the figure as well. Using more images usually benefits the performance of AFV and FFV on both datasets. In specific, we can see a dramatic performance boost with the number of images increased from 5 to 40 per class for UCF101. A further increase of images does not improve the performance significantly, which is especially true in the inductive setting. For HMDB51 dataset, the similar trend of performance improvement can be observed from Fig.\ref{fig_noImgs}, and the performance improvement stops until the number of images per class increases to around 80. In addition, the proposed image-based semantic representations using only 5 images per class can achieve better performance on UCF101 than attributes and word vectors, and the number rises to 20 for HMDB51 to beat word vectors. To summarise, we are able to use a small number of relevant images to encode the semantic representations of human actions, yet boosting the zero-shot human action recognition accuracy to a large extent.
\section{Conclusions and Future Work}
We explore the alternative side information to the existing attributes and word vectors towards improved zero-shot human action recognition. The textual descriptions of human actions from the Internet can be used as side information for knowledge transfer in ZSL. In addition, the combination with pre-trained word vectors can further improve the power of text-based semantic representations, even better than the manually annotated attributes. On the other hand, the image-based semantic representations achieve dramatic performance improvement compared with the ones based on other side information (e.g., texts and human annotations), due to the narrower semantic gap. Our experiments also show that a small number of images are enough to gain significant performance improvement.

There are quite a few directions we can follow in our future work. Firstly, we only use a very simple encoding method (TD) for text-based semantic representations in this paper, which results in an extremely high dimensionality and sparse vector representation per document. It has been chosen in this work as a proof of concept, but could be optimised by using alternative techniques such as latent Dirichlet allocation (LDA), latent semantic indexing (LSI), etc. Besides, in our methods of text-based representation encoding, only the occurrences of different words in a given document are considered, and the word orders which play an important role in text understanding have been ignored. Thus the meaning of sentences containing ``not" and ``but" would be destroyed. To overcome this limitation, some potential techniques recently developed in NLP (e.g., \textit{document2vec} \cite{le2014distributed}) would be investigated. Currently, we extract image features with deep CNN models pre-trained on large scale object classification dataset (i.e., ImageNet). Although the pre-trained models have already shown great generalization and transferability to other visual recognition tasks, better performance can be expected by fine-tuning the models with our specific human action image data. We have done some preliminary experiments on the combination of two different types of semantic representations, but only get results no better than the use of image-based semantic representation alone. We do not want to rush to the conclusion that the image- and text-based semantic representations are not complementary before further studying the combination methods in our future work.
\subsubsection*{Acknowledgments.} The authors would like to thank Ubai Sandouk from MLO group at The University of Manchester for personal communication and the anonymous reviewers for their valuable comments and suggestions.

%\section{The References Section}\label{references}
\bibliographystyle{splncs03}
\bibliography{ecml2017bib}

\end{document}